# Long-Term Hourly Scenario Generation for Correlated Wind and Solar Power combining Variational Autoencoders with Radial Basis Function Kernels

Julio A. S. Dias (alberto@psr-inc.com)

PSR Energy Consulting

*Abstract* — **Accurate generation of realistic future scenarios of renewable energy generation is crucial for long-term planning and operation of electrical systems, especially considering the increasing focus on sustainable energy and the growing penetration of renewable generation in energy matrices. These predictions enable power system operators and energy planners to effectively manage the variability and intermittency associated with renewable generation, allowing for better grid stability, improved energy management, and enhanced decision-making processes. In this paper, we propose an innovative method for generating long-term hourly scenarios for wind and solar power generation, taking into consideration the correlation between these two energy sources. To achieve this, we combine the capabilities of a Variational Autoencoder (VAE) with the additional benefits of incorporating the Radial Basis Function (RBF) kernel in our artificial neural network architecture. By incorporating them, we aim to obtain a latent space with improved regularization properties.**

**To evaluate the effectiveness of our proposed method, we conduct experiments in a representative study scenario, utilizing real-world wind and solar power generation data from the Brazil system. We compare the scenarios generated by our model with the observed data and with other sets of scenarios produced by a conventional VAE architecture. Our experimental results demonstrate that the proposed method can generate long-term hourly scenarios for wind and solar power generation that are highly correlated, accurately capturing the temporal and spatial characteristics of these energy sources. Taking advantage of the benefits of RBF in obtaining a well-regularized latent space, our approach offers improved accuracy and robustness in generating long-term hourly scenarios for renewable energy generation.**

**We anticipate that the application of our proposed method can significantly contribute to the efficient planning and operation of electrical systems, enabling better integration of large-scale wind and solar energy and facilitating the transition towards a sustainable energy future.**

## I. INTRODUCTION

The increasing demand for renewable energy sources, such as wind and solar power, has posed significant challenges for the planning and operation of electrical systems. Accurate generation of realistic future scenarios of renewable energy generation is crucial to ensure grid stability and to facilitate optimal decision-making of resource allocation. Traditional methods for generating long-term scenarios often rely on statistical techniques or simplistic assumptions, which may not capture the complex temporal and spatial correlations inherent in renewable energy data, mainly due to the need to model this type of generation using hourly discretization. As a result, there is a need for innovative approaches that can generate realistic and correlated long-term hourly scenarios for wind and solar power generation.

On one hand, Variational Autoencoders (VAEs) have emerged as a promising tool for modeling complex data distributions and generating realistic samples. VAEs are a type of generative model that can learn the underlying structure of high-dimensional data and generate new samples from that distribution. By combining neural networks and probabilistic modeling, VAEs can capture the latent space representation of the input data, enabling the generation of diverse and coherent samples. On another hand, although VAEs have shown remarkable success in generative modeling, there are several drawbacks and limitations when applied to time series, particularly in the context of multivariate time series with many variables to be simultaneously generated.

The objective of this paper is to investigate the application of a proposed extended architecture of VAEs in generating long-term hourly scenarios for wind and solar power generation that incorporates Radial Basis Function (RBF) kernel to improve the regularization properties of the latent space. This enhancement aims to ensure a better representation of the underlying data distribution and increase the reliability and accuracy of the generated samples. Through this extended architecture, we seek to provide a more robust and effective approach for generating realistic and correlated scenarios of wind and solar power generation, which can contribute to improved long-term planning and decision-making in the context of sustainable energy systems.

The basic approach proposed in this paper is that the scenarios can be generated on a weekly basis and then disaggregated on an hourly basis using a disaggregation profile selection mechanism that assesses the similarity of the synthetic scenario with the historical data. For this purpose, the inherent structure of VAEs, that were originally designed for semi supervised learning, allows for the training of an architecture that can not only generate scenarios but also identify the best profiles from historical data, given that the latent space constructed by VAEs can be interpreted as a space that measures similarities among the samples.

We applied the proposed architecture to the Brazilian power system, which consists of hundreds of wind and solar power plants, analyzing the adherence of the generated scenarios to historical data and comparing the results with other set of scenarios generated by a conventional VAE architecture. The preliminaries results highlight the advantages of our approach in terms of improved accuracy and representation of renewable energy generation.

## II. VARIATIONAL AUTOENCODER

Variational Autoencoders (VAEs)[1] have gained significant attention as powerful generative models capable of capturing complex data distributions. At the core of a VAE is the encoder network, which takes an input data point and maps it to a latent space representation. The encoder network typically consists of multiple layers of neural networks, such as convolutional or fully connected layers. These layers encode the input data into a lower-dimensional representation, capturing its essential features and reducing dimensionality.

The latent space serves as the bottleneck of the VAE architecture, where the encoder maps the input data. It is a lower-dimensional space that captures the underlying distribution of the data. In a VAE, the latent space is assumed to follow a multivariate gaussian distribution, characterized by a mean vector ($\mu$) and a diagonal covariance matrix ($\sigma$). This distribution allows for the generation of diverse samples in the latent space.

The decoder network plays a crucial role in a VAE as it reconstructs the data from the latent space representation. It takes a sample from the latent space as input and generates a reconstruction that closely resembles the original input. Like the encoder network, the decoder network consists of several layers, often mirroring the structure of the encoder network but in reverse order. The decoder network aims to learn the mapping from the latent space back to the original data space, reconstructing the input data with high fidelity.

A conceptual VAE architecture is presented in figure 1.

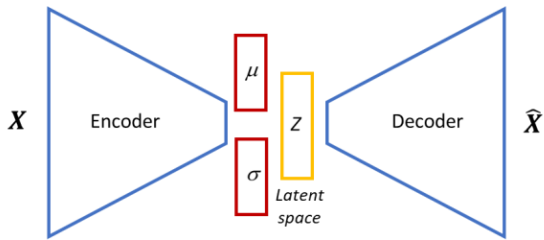

Figure 1. VAE architecture

The "reparameterization trick" is a fundamental idea of VAEs because it enables the training of the neural network architecture while maintaining differentiability of the neurons. It addresses the challenge of sampling from the latent space, while allowing for backpropagation and gradient-based optimization. Instead of directly sampling from the latent space distribution, the trick involves sampling from a standard gaussian distribution and then applying a linear transformation based on the mean and covariance parameters learned by the encoder network. By decoupling the sampling process from the network parameters, the reparameterization trick ensures that the network remains differentiable, enabling the learning of meaningful representations in the latent space.

The topology of the "reparameterization trick" is presented in figure 2.

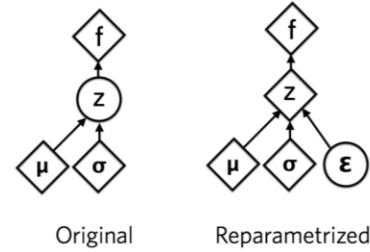

Figure 2. The reparameterization trick

The nodes corresponding to the mean and standard deviation are separated into deterministic nodes, while an additional stochastic node is introduced ($\varepsilon$). This stochastic node is responsible for generating a random sample from a standard gaussian distribution, $\varepsilon \simeq N(0,1)$, which is then multiplied by the standard deviation vector and added to the mean vector. This design choice facilitates the training process through backpropagation, as the gradients can be efficiently computed and used to update the network parameters given that the training process only iterates over the deterministic nodes.

The training of a VAE is driven by a loss function that encourages the latent space to capture the underlying data distribution and generate accurate reconstructions. The loss function consists of two components: the reconstruction loss and the regularization loss. The reconstruction loss measures the difference between the input and the output of the decoder network, while the regularization loss encourages the latent space to follow a desired distribution, typically a gaussian distribution. The regularization loss is commonly expressed as the Kullback-Leibler (KL) divergence [2] between the learned latent space distribution and the standard gaussian distribution.

The reconstruction loss is often given by the mean squared error (squared L2 norm) between each element in the encoded $X$ and the decoded $\hat{X}$:

$$L_{recons} = l(X, \hat{X}) = \{l_1, \cdots, l_N\}^T \qquad (1)$$
$$l_n = (X_n - \hat{X}_n)^2$$

While the KL loss is given by:

$$L_{KL} = -0.5 \sum 1 + log(\sigma^2) - \mu^2 - \sigma^2 \qquad (2)$$

The overall loss is given by:

$$L = L_{recons} + L_{KL} \qquad (3)$$

During training, random batches of data are fed into the VAE, and the gradients are computed to update the network parameters. The optimization process aims to minimize the overall loss function through backpropagation and gradient descent. The training iterates multiple times until the model converges, capturing the underlying data distribution in the latent space and enabling the generation of new samples.

While VAEs have shown remarkable success in generative modeling, there are several drawbacks and limitations when applied to time series generation [3], [4], particularly in the context of multivariate time series with many variables to be simultaneously generated. One of the main challenges arises from the difficulty in capturing long-term dependencies and complex temporal patterns present in such multivariate time series data. The latent space may not adequately represent the intricate dynamics and interdependencies among the variables over time, leading to suboptimal generation results. Moreover, VAEs are prone to producing smoothed or blurry outputs[5],[6], which becomes even more pronounced when dealing with multivariate time series. The trade-off between reconstruction accuracy and the regularization term in the loss function can lead to a loss of fine-grained details and high-frequency components, resulting in less accurate and less realistic generated sequences. Additionally, the issue of mode collapse [6], where the generated samples become overly similar and lack diversity, can be exacerbated in the multivariate setting, further limiting the variety of generated patterns. Furthermore, the requirement for a large amount of training data becomes even more crucial when dealing with multivariate time series, as capturing the complex interactions between multiple variables necessitates a richer dataset.

## III.  PROPOSED MODEL

In this section, we present our proposed model, which aims to overcome the limitations and drawbacks of the original VAE architecture, for generating long-term hourly scenarios of renewable energy generation. The core ideas of our model consist of three main components, designed to address these challenges and improve the accuracy and reliability of the generated scenarios.

Firstly, we separate the generation process into two distinct modules: the mean weekly scenario generation module and the hourly scenario disaggregation module. This separation allows us to capture both the macroscopic trends of the weekly average scenario and the fine-grained variations at the hourly level.

Secondly, to improve the modeling capabilities of the VAE, we introduce a modified architecture that incorporates the Radial Basis Function (RBF) kernel as an integral part of its encoder/decoder framework.

Lastly, to address the challenge of selecting appropriate disaggregation profiles for the hourly scenarios, we leverage the VAE clustering characteristics and the features extracted from the latent space. This allows us to construct a selection function that identifies and selects the most suitable profiles for accurately representing the hourly variation in renewable energy generation given the sampled week scenario.

### A.  First Module

To introduce the first module of the proposed model, it is initially necessary to present the Radial Basis Function (RBF) kernel, which is a popular mathematical function used in machine learning and data analysis. It is commonly employed in various applications, including support vector machines, clustering algorithms, and dimensionality reduction techniques [7]. The RBF kernel calculates the similarity or distance between data points based on their radial distance from each other in a multi-dimensional feature space. It is characterized by its flexibility in capturing complex and non-linear relationships within the data. It assigns higher weights to points that are closer together and gradually decreases the weights as the distance between points increases. This allows the RBF kernel to capture both local and global patterns in the data. The kernel's ability to effectively model complex data distributions makes it particularly useful in applications where data points are not easily separable or exhibit non-linear relationships.

The formula for the RBF kernel is:

$$K(x,y) = exp(-\gamma \|x-y\|^2) \qquad (4)$$

Where:
$K(x,y)$ represents the value of the kernel between points $x$ and $y$;
$\|x-y\|^2$ is the squared Euclidean distance between points $x$ and $y$;
$\gamma$ is a parameter that controls the influence of the distance

The encoder layer of an autoencoder is designed to transform the input data into a compressed representation in the latent space. This compressed representation captures the most salient features of the input data. The architecture of the encoder layer often consists of multiple hidden layers, each comprising a set of neurons or nodes that perform nonlinear transformations on the input data. These transformations gradually reduce the dimensionality of the data, extracting higher-level features and representations as we move deeper into the network.

In figure 3 is presented a typical architecture of an encoder layer.

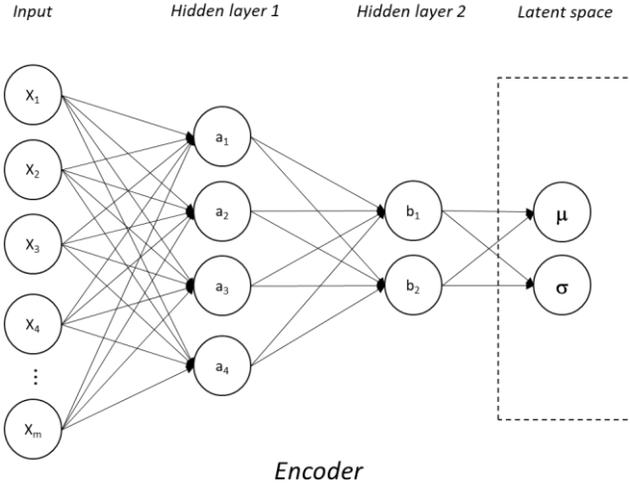

Figure 3. VAE – Encoder layer

For example, in an image autoencoder, the encoder layer may comprise a series of convolutional layers followed by fully connected layers. The convolutional layers apply filters to extract local patterns and spatial information from the image, while the fully connected layers further process the extracted features and compress them into a lower-dimensional representation [8]. The size of the latent space, or bottleneck layer, in the encoder is a key parameter that determines the level of compression and the information capacity of the autoencoder. A smaller latent space size results in more aggressive compression, while a larger size preserves more information but may reduce the efficiency of the encoding process.

In our proposed architecture, presented in figure 4, the first layer is replaced by the RBF kernel layer.

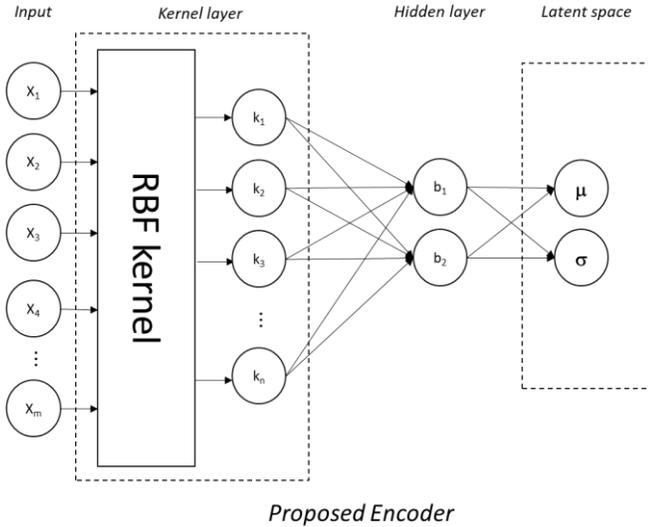

Figure 4. Proposed VAE architecture – Encoder layer

The addition of the RBF kernel layer facilitates the extraction of meaningful and informative features right from the first layer. The regularization properties of the kernel, along with its ability to capture complex patterns and dependencies in the data [7], enable the encoder to learn a more organized and structured representation of the input. By establishing a linear relationship between the features in the transformed space generated by the RBF kernel layer, this transformation allows for subsequent encoder layers to become simpler and more concise compared to architectures required to encode raw inputs. This reduction in complexity is achieved because the transformed space already encapsulates important features and relationships of the data. The subsequent layers can focus on refining and disentangling the underlying factors of variation in a more straightforward manner, leading to a more efficient and streamlined modeling of the data distribution.

This reduction in complexity of the hidden layers enables a less intricate transformation to the latent space without sacrificing the efficiency of the encoder process.

Incorporating the RBF kernel layer into the encoder of a VAE introduces additional advantages, particularly in terms of computational efficiency during the training process. One key advantage is that the RBF kernel has a single hyperparameter, gamma, which is not optimized along with the neural network training. This characteristic allows for pre-calculating the outputs of the kernel layer for a specified gamma, significantly reducing the computational cost during the network training. By pre-calculating the outputs of the RBF kernel layer, we can effectively bypass the need for repetitive computations in each training iteration. This optimization greatly reduces the overall computational burden and accelerates the training process of the VAE. The precalculated outputs of the kernel layer can be stored and reused throughout the training phase, eliminating the need for re-computation during each forward and backward pass of the network.

This computational efficiency is particularly valuable when dealing with large datasets containing hundreds or thousands of renewable power plants.

Finally, the decoder architecture of our proposed VAE model, presented in figure 5. consists of intermediate hidden layers followed by a final layer responsible for applying a transformation that corresponds to the inverse function of the radial basis function (RBF) kernel. This design enables the reconstruction of the output in the domain of the original variables.

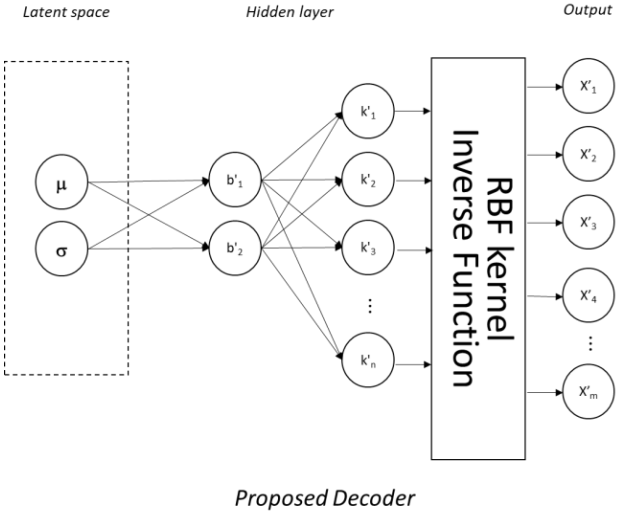

*Proposed Decoder*

Figure 5. Proposed VAE architecture – Decoder layer

Unlike the direct application of the RBF kernel transformation, applied in the encoder layer, which has a parametric formula, constructing the inverse transformation is not a trivial task. In this work, we explore two different alternatives to build this layer. In the first alternative, we propose not to explicitly define the inverse transformation but rather implicitly map it by introducing additional intermediate layers. By doing so, we allow the optimizer to identify the optimal parameters for achieving a good reconstruction of the original data. In the second alternative, we build a separate neural network whose sole purpose is to learn the inverse mapping of the RBF kernel. This second neural network is trained independently of the scenario generation task. Once trained, its structure is integrated into the final architecture of the decoder. This way, the optimization process of the proposed VAE framework can find the best intermediate layers that connect to this sub-network.

## B. Second Module

The second module of our proposed model utilizes the latent space constructed in the VAE of the first module as a function to measure similarity. This module aims to exploit the encoded representations in the latent space to determine the observation from the original dataset that exhibits the closest resemblance to the generated sample. By utilizing the neural network and the clustering characteristics of the latent space, we can develop a selection function that identifies the most suitable profiles for disaggregating the weekly average scenario into hourly values.

To facilitate the visualization of the scheme, figure 6 provides a simplified visual representation of the mapping process in the latent space. On the left-hand side, a 2D graph illustrates the relationship between weekly discretized solar and renewable energy generation, showcasing three example vectors. The axes represent the respective energy generation values. Adjacent to this graph, on the right-hand side, a unidimensional latent space is depicted, with the three sample vectors mapped onto it.

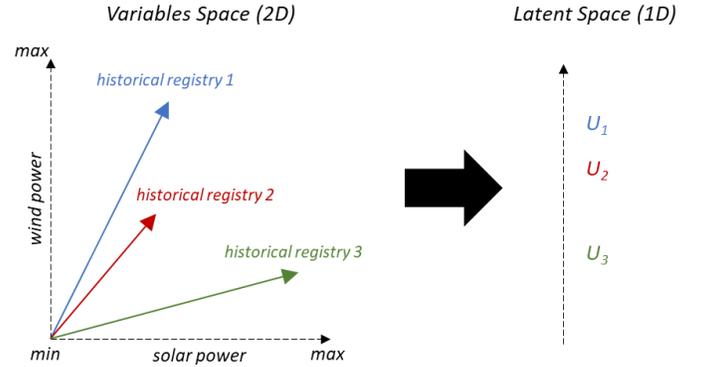

Figure 6. Example of Space Transformation

Figure 7 demonstrates the process of generating a scenario from this latent space. On the left-hand side of the figure, a sample is generated in the unidimensional latent ($S_1$). This sample is then mapped back into the variable space, shown on the right-hand side of the figure.

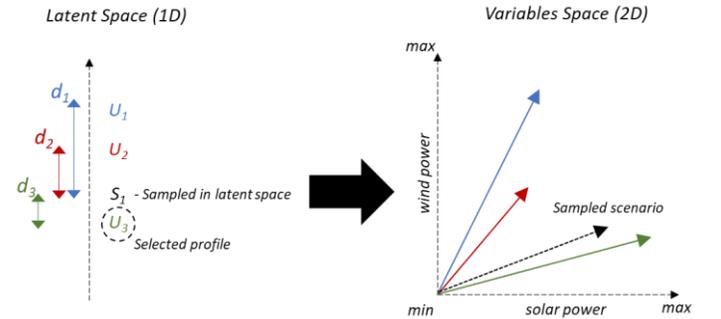

Figure 7. Example of Scenario Generation

In this example, as $d_3 < d_1 < d_2$, the historical data associated with $U_3$, discretized at an hourly level, is selected to disaggregate the newly generated scenario.

In practice, in a VAE, the mapping from the variable space to the latent space does not correspond to a direct mapping to a single point in the latent space, but rather to a distribution. Therefore, the selection function for profile disaggregation needs to consider this characteristic when measuring similarity.

To address this issue, we propose the use of Mahalanobis distance as a measure of similarity. Mahalanobis distance is a statistical measure that considers both the covariance structure and the mean values of variables. It provides a way to calculate the distance between points and distributions, taking into consideration the variability present in the data [9].

The general formula for the Mahalanobis distance in the multidimensional case is given by:

$$D^2 = (z' - \mu_n)^T \Sigma_n^{-1} (z' - \mu_n) \quad (5)$$

Where:

$D$ is the Mahalanobis distance;

$z'$ is the vector representing a sampled point on latent space;

$\mu_n$, $\Sigma_n^{-1}$ characterizes the distribution of the $n$-th observation on latent space;

Additionally, given that in VAEs, the latent space is optimized in a way that the covariance matrix ($\Sigma$) becomes diagonal. This means that the Mahalanobis distance formula can be simplified, as the diagonal matrix only contains variances and no cross-covariance terms.

Figure 8 illustrates an example of the disaggregation of historical data for a solar plant (shown in the upper part of the graph) and a wind plant (shown in the lower part of the graph). The red line represents the aggregated historical data for each plant, already averaged on a weekly basis, and remains constant within each week. The subplots appearing below each hourly generation series correspond to the extracted profiles.

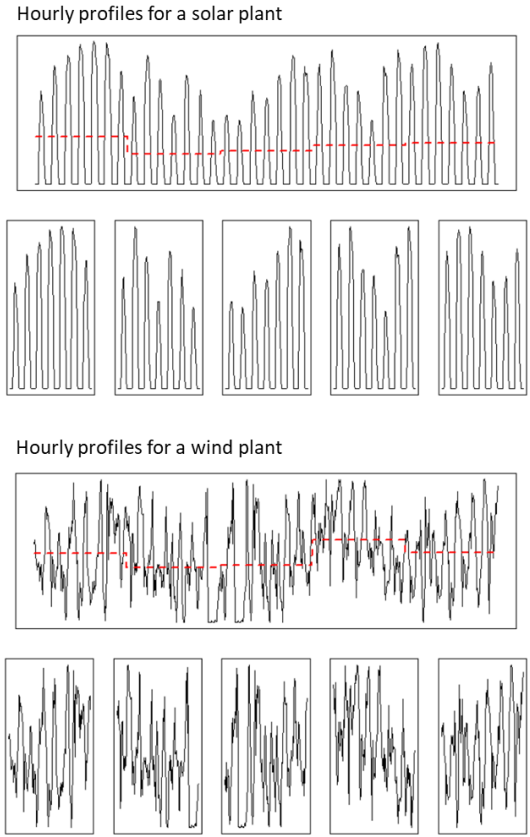

Figure 8. Historical data disaggregation examples

Later, during the scenario generation process, these profiles will be chosen according to the similarity measured between the weekly scenario generated and the set of weeks of the historical data.

## C. Consolidated Architecture

Consolidating the points discussed, the steps for estimation and simulation of the proposed architecture for generating hourly scenarios are presented in the following procedures:

The first procedure outlines the estimation phase of the proposed model:

1. Construct a dataset of renewable energy generation observations (wind/solar) with weekly discretization by aggregating the raw hourly data into weekly data points.
2. Extract historical profiles of hourly generation based on the relationship between hourly and weekly data.
3. Split the data into training and test sets.
4. Define the first layer of the encoder, which incorporates the RBF kernel with a specified gamma value.
5. Define the intermediate hidden layers of the encoder, consisting of neurons with activation functions such as ReLU, Sigmoid, etc.
6. Incorporate the specific components of the VAE architecture, such as the reparametrization trick, and the loss function.
7. Define the hidden intermediate layers of the decoder architecture, also comprising neurons with activation functions such as ReLU, Sigmoid, etc.
8. Incorporate the inverse transformation layer of the RBF kernel into the decoder architecture to enable the reconstruction of the input space.
9. Train the designed neural network by optimizing the parameters using appropriate optimization techniques like backpropagation and stochastic gradient descent.
10. Validate the results for the test set using different gamma values and select the model that provides the best performance metrics.

The second procedure outlines the scenario generation phase (per week/scenario) of the proposed model:

1. Generate a random sample of epsilon from $\varepsilon \simeq N(0,1)$.
2. Calculate the equivalent sample in the latent space.
3. Feed the decoder structure with the generated sample and obtain the resulting generation scenario (weekly discretization).
4. Calculate the Mahalanobis distances of the sample in the latent space with respect to the distributions obtained during the training phase and determine the index of the reference historical observation that exhibits the highest similarity to the generated scenario.

5. Apply the hourly profile associated with this historical observation to the sampled weekly scenario to produce the final hourly scenario.

By following these steps, the capabilities of the proposed VAE architecture and the disaggregation scheme ensure the generation of realistic and representative hourly scenarios.

## IV. RESULTS

In this section, we present the application of our proposed model to generate hourly scenarios for a dataset comprising 145 renewable energy power plants in Brazil, including both wind and solar plants. The evaluated plants are shown in figure 9 (yellow marks represent solar plants, blue marks represent wind plants and green circles represent aggregation of multiple plants). It was generated a set of 1000 hourly scenarios with a 5-year horizon using the developed architecture, because this particular horizon of study is widely applied in the operational planning of the Brazilian power system.

The objective of this study was to assess the effectiveness and performance of our model in capturing the temporal and spatial characteristics of renewable energy hourly generation.

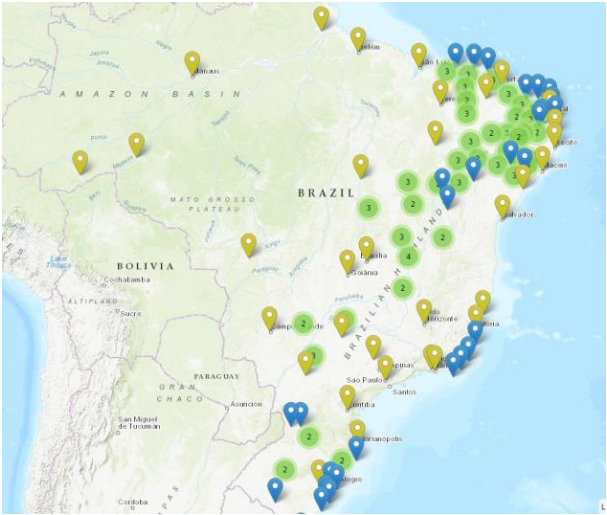

Figure 9. Location of the Evaluated Power Plants in Brazil [10]

In our analysis, we explored two different alternatives for the decoder architecture in our proposed model. One approach involved constructing the decoder with an explicit, pre-trained topology that represents the inverse function of the RBF kernel, attached to the decoder hidden layers. Another approach involved defining the inverse function implicitly through the composition of the decoder's layers.

Figure 10 shows the evolution of the loss function during the training of the two alternative neural network architectures: one with the explicit definition of the inverse RBF kernel function, pre-trained separately, and the other with the implicit definition through additional final layers in the VAE decoder. Comparing the plots, we observe that although the architecture with the explicit definition initially shows faster decay of the loss function, after a few epochs, the performance of the network with the implicit definition becomes better, stabilizing around a significantly lower minimum loss function value. This indicates that the training process of the network with the implicit definition converges more quickly and efficiently. As a result, for the same maximum number of epochs, the implicit architecture produces better scenarios.

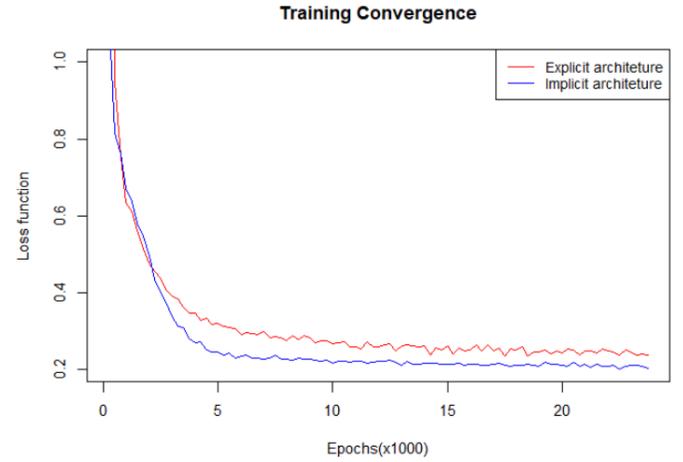

Figure 10. Loss Function During the Training Process

Additionally, it is noteworthy that throughout the experiments and evaluation of different parameterizations, it was observed that the results of the implicit architecture remained more robust and constant than the architecture with explicit representation, which showed a greater variation according to the adopted parameters. For this reason, the results adopted in this section refer to those obtained with the implicit architecture.

To comprehensively analyze the results, we propose a combination of qualitative graphical analysis, which enables a quick visual assimilation of the model's performance, with quantitative numerical analysis using robust statistical tests appropriate for the specific analysis objective. This approach allows for a more holistic evaluation of the model's performance, considering both visual patterns and statistical measures. By integrating these two types of analysis, we gain a deeper understanding of the model's effectiveness in capturing the underlying dynamics of the renewable energy generation dataset.

Firstly, to provide a better qualitative analysis of the model's ability to capture various statistical patterns in the data, figure 11 presents examples of marginal probability density functions for some power plants. These plants exhibit diverse shapes and asymmetries in their probability distributions. By comparing the densities obtained from historical data with those obtained from the generated scenarios, it is observed that the model successfully reconstructs the distinct distribution patterns. This observation highlights the model's capacity to capture the inherent characteristics and statistical properties of the renewable energy generation, thereby demonstrating its effectiveness in generating scenarios that accurately represent the underlying data distribution.

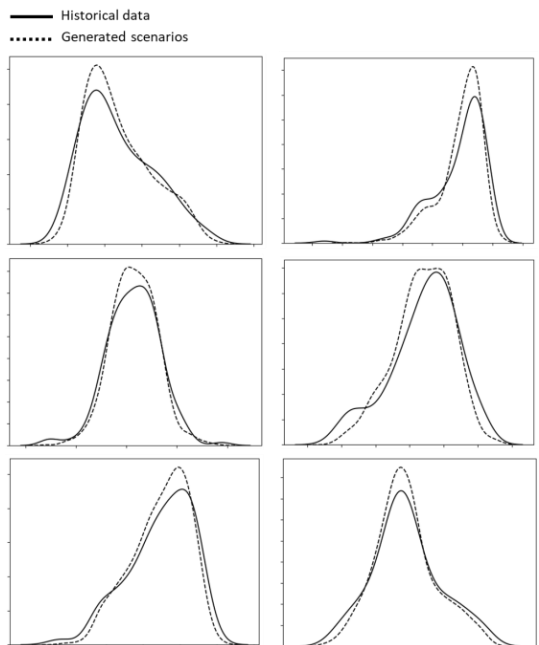

Figure 11. Probability Density Functions for Some Power Plants

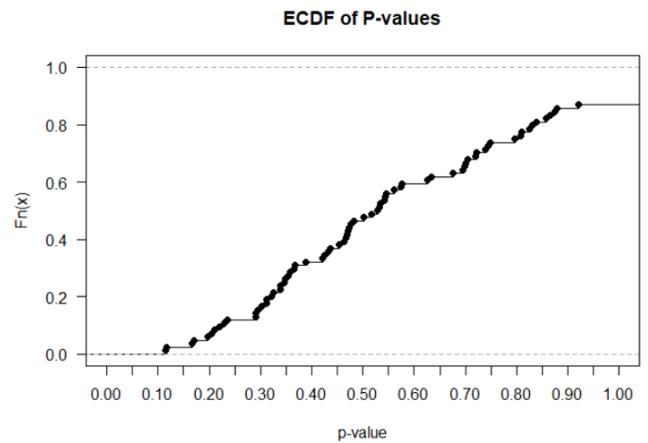

Figure 12. Cumulative distribution function of all p-values

In order to perform now a quantitative analysis of the marginal distributions obtained, we compare if two different samples have similar marginal probability distributions using the statistical test called Kolmogorov-Smirnov (KS) test [11]. The KS test compares the empirical cumulative distribution functions (CDFs) of the two samples and assesses whether they are statistically different. The result of the KS test will include the p-value, which indicates the probability of obtaining a value as extreme or more extreme than the observed, assuming that the two samples are from the same distribution. When interpreting the test result, if the p-value is greater than or chosen significance level (0.05), we fail to reject the null hypothesis that the two samples have similar marginal probability distributions. Otherwise, if the p-value is less than the significance level, we reject the null hypothesis and conclude that the two samples have different distributions.

Remarkably, all the results exhibited p-values above 0.05, indicating that the synthetic scenarios generated by our model are statistically indistinguishable from the historical data. To provide a comprehensive overview of these results, we present the cumulative distribution of the p-values obtained for all power plants in figure 12. The high p-values across all the tests further reinforce the model's ability to accurately reproduce the statistical characteristics of the original data.

Analogous to the qualitative comparison of marginal distributions, we now present visual comparative analysis of correlations.

In figure 13 is shown contour plots comparison of three different combinations of renewable plants. These contour plots provide a visual representation of the joint probability distribution between two variables. By observing these plots, we can qualitatively assess how well the 2D densities of the generated scenarios from our proposed model align with the 2D densities of the historical data. This observation indicates that the proposed model effectively captures and reproduces the joint density patterns present in the historical data, demonstrating its ability to generate scenarios that maintain similar statistical characteristics.

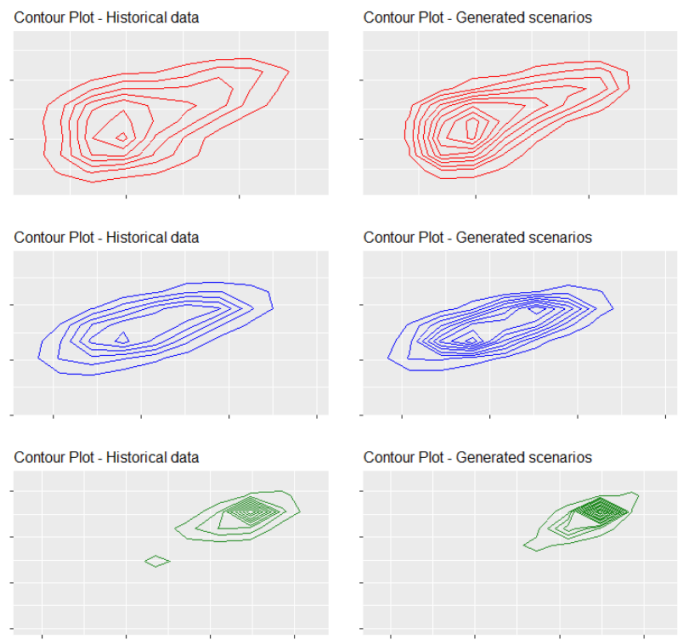

Figure 13. Contour plots Comparison

In figure 14 is shown a correlation heatmap. To facilitate visualization, we consider only a subset of 20 power plants in the spatial correlation heatmap graph. The heatmap illustrates the correlation matrix obtained from historical data and the correlation matrix derived from the synthetic scenarios, focusing on this subset of power plants. Notably, the model exhibits a remarkable ability to capture the intricate correlation patterns present in the historical data.

Visually, it is difficult to discern any noticeable differences between the correlation patterns of the historical data and the generated scenarios, further validating (albeit qualitatively) the model's effectiveness in capturing and reproducing the complex interdependencies between variables.

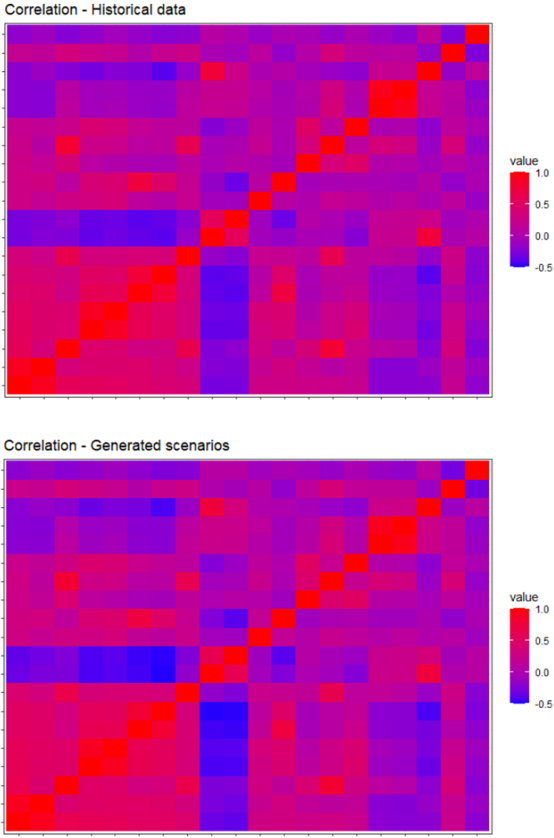

Figure 14. Correlation Matrix Heatmaps

Comparing the spatial correlations of each combination of all plants, the mean absolute error between the estimated and historical correlation is only 0.034 while the maximum is 0.15. In figure 15 is presented the histogram of all absolute errors calculated.

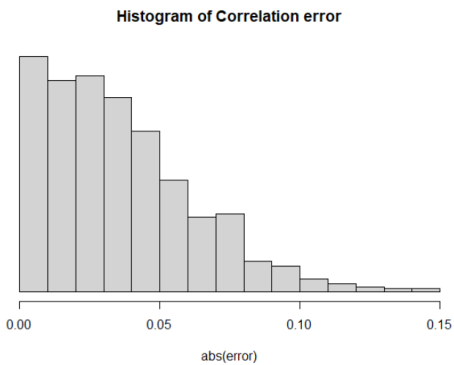

Figure 15. Histogram of Correlation Error

In figure 16 it is shown the disaggregated hourly scenarios generated for a complete week. The upper part of the graph displays the scenarios for a solar plant, while the lower part shows the scenarios for a wind plant. For each plant, two subplots are presented. In the first subplot, three scenarios are overlaid on the historical hourly generation average for each plant. This allows us to observe the evolution of each scenario over time and compare it to the historical average. In the second subplot, all the generated scenarios are presented together, overlaid on the historical hourly average. This visualization provides an overview of the dispersion of the scenarios around the average.

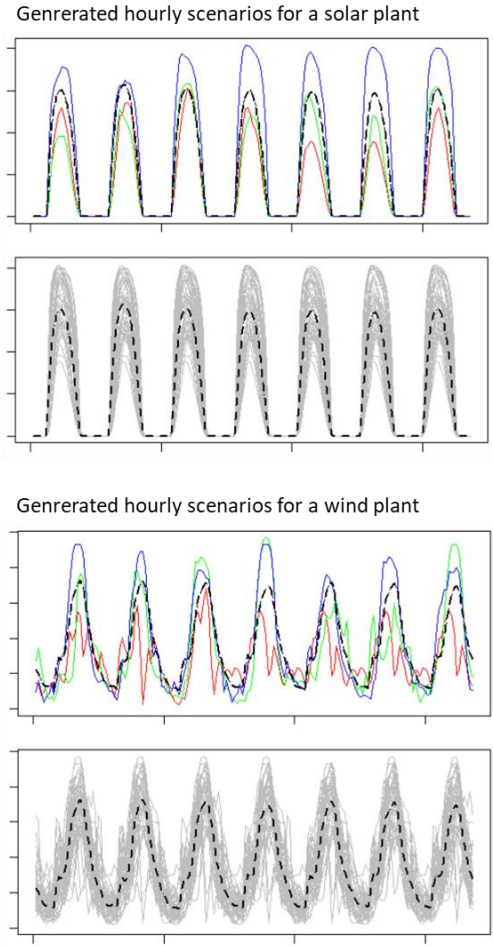

Figure 16. Hourly Scenarios Generated for Some Power Plants

In the following part, we will present some results related to the traditional VAE architecture (named pure VAE) to compare the benefits of the proposed model. However, it is important to note that the same strategy proposed for hourly disaggregation in the proposed model will be applied to the VAE architecture. Therefore, the only aspect being analyzed is the benefit of introducing the RBF kernel layer. It is expected that if the traditional VAE architecture were directly applied to the hourly time series, the results would be even more disparate.

In figure 17, it is shown a graph depicting the p-values obtained from statistical tests that assess the representation capacity of the marginal probability density functions in the pure VAE architecture. This graph is similar to the one added for the proposed model, allowing for a comparative analysis between the two architectures.

It is observed that the majority of renewable energy plants in the pure VAE architecture still pass the statistical test, indicating that their marginal probability density functions are well-represented. However, it is worth noting that, unlike the proposed model, approximately 20% of the plants do not pass the test, for the same chosen significance level (0.05), suggesting that the pure VAE model falls short in accurately capturing the probability distributions for these specific plants.

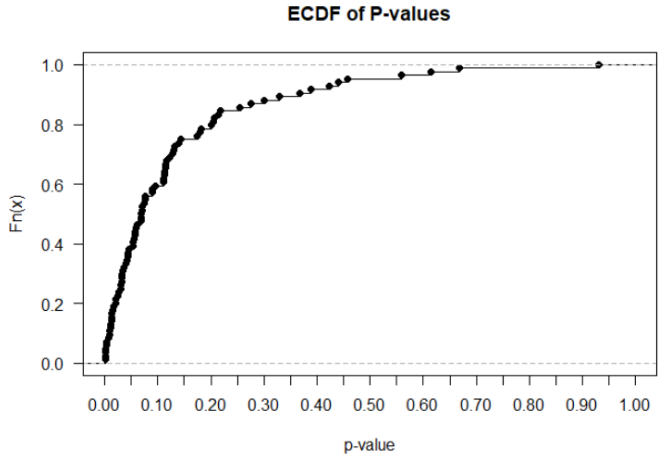

Figure 17. Cumulative distribution function of all p-values (pure VAE)

Figure 18 shows two comparisons of marginal probability densities between the historical data, pure VAE architecture and the proposed model. At the top, the estimated density functions are presented for a renewable plant that is accepted in the statistical test, while at the bottom, the probability density functions are presented for a plant that is rejected in the statistical test (for pure VAE architecture).

It is observed that for the first renewable energy plant, the proposed model does not bring significant improvements in the representation compared to the VAE architecture. The VAE architecture already provides a satisfactory representation of the probability density for this first plant. However, for the second renewable energy plant, the proposed model demonstrates significant enhancements in representing the probability density.

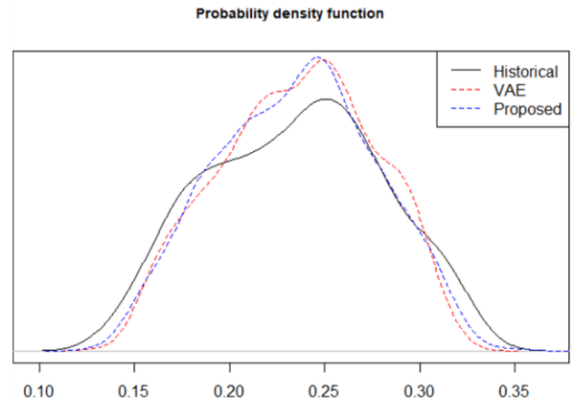

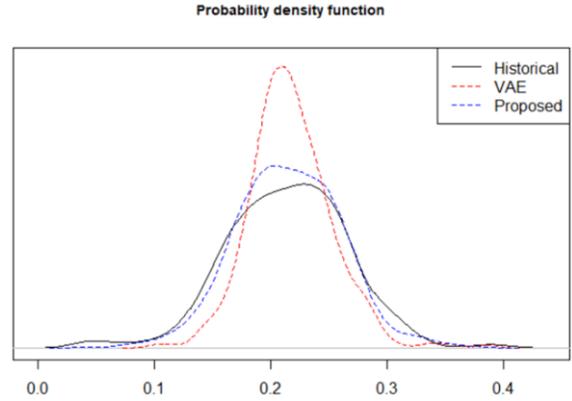

Figure 18. Probability Density Functions for two Power Plants

The same comparison of correlations performed for the proposed model was also conducted for the pure VAE architecture. In this case, it was observed that the mean absolute error practically doubled, increasing from 0.034 to 0.0642. Furthermore, in figure 19, two X-Y graphs are presented. The X-axis represents the observed correlation in the historical data, while the Y-axis represents the correlation in the VAE architecture, on the left side, and the correlation in the proposed model, on the right side.

Ideally, the points should fall on the dotted line in the graph, and thus, the closer they are to the line, the better.

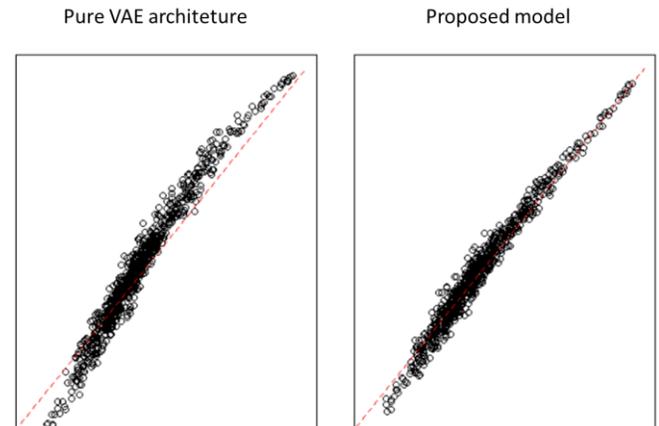

Figure 19. Plot X-Y of Correlations

It is noteworthy that for the proposed model, the spatial correlations tend to align more closely compared to the conventional VAE architecture, particularly for the larger positive and negative correlations. This indicates that the proposed model performs better in capturing and reproducing correlations, resulting in a more accurate representation of the underlying relationships between the renewable energy plants.

## V. CONCLUSION

In this study, we proposed a novel approach for generating hourly scenarios of renewable energy generation using a two-module model. The first module consisted of a modified VAE architecture that incorporated a RBF kernel in its encoder and decoder structure. This allowed for the extraction of meaningful and informative features from the data, promoting more regularity in the transformed space. As a result, the subsequent layers of the encoder became simpler and more concise, facilitating the modeling of the data distribution. The second module utilized the latent space constructed in the VAE as a basis for measuring similarity. By exploring the characteristics of the latent space, a selection function was established to identify the most appropriate profiles for accurately disaggregating the weekly average scenario into hourly values. The Mahalanobis distance was employed as a measure of similarity, taking into account the distributional nature of the latent space.

To evaluate the performance of our proposed model, we applied it to the task of generating hourly scenarios for a dataset of 145 renewable energy plants in Brazil, including both wind and solar plants and the results demonstrated the effectiveness of our proposed model in generating accurate and realistic hourly scenarios of renewable energy generation.

We also compared different decoder architectures, and discussed the potential benefits associated with each approach.

Future work could explore further improvements to the model, such as investigating alternative kernels and different parametrizations for the explicit inverse function.

Future work could also explore the adaptation of the proposed model for short-term forecasting tasks. The model could be fine-tuned to predict renewable energy generation for the next day or next week. This extension would enable the utilization of the model as a tool for short-term forecasting, providing valuable insights for energy system operators and facilitating decision-making in real-time scenarios. Additionally, the integration of external factors such as weather forecasts could further enhance the predictive capabilities of the model. By considering these aspects, the proposed framework could be applied to a wider range of applications, encompassing both long-term scenario generation and short-term forecasting in the renewable energy domain.

Furthermore, the proposed framework could be extended to other renewable energy sources and geographical regions, expanding its applicability, and providing valuable insights for a broader range of energy systems.